\documentclass[11pt,a4paper]{article}
\usepackage[hyperref]{acl2020}
\usepackage{times}
\usepackage{latexsym}

\usepackage{microtype}

\usepackage{amsmath}
\usepackage{amssymb}
\usepackage[outline]{contour}
\usepackage{multirow,makecell}
\usepackage{graphicx}
\usepackage[utf8]{inputenc}
\usepackage{booktabs}
\usepackage{siunitx}
\usepackage{tree-dvips}
\usepackage{tikz}
\usetikzlibrary{calc}
\usepackage{tikz-qtree}
\usepackage{adjustbox}
\usepackage{subcaption}
\usepackage[shortlabels]{enumitem}

\usepackage{url}

\usepackage{algorithm}
\usepackage{algpseudocode}

\newcommand{\xl}[0]{\rotatebox[origin=c]{45}{$\rightarrow$}}
\newcommand{\xr}[0]{\rotatebox[origin=c]{-45}{$\leftarrow$}}
\newcommand{\xL}[0]{\rotatebox[origin=c]{45}{$\Rightarrow$}}
\newcommand{\xR}[0]{\rotatebox[origin=c]{-45}{$\Leftarrow$}}

\aclfinalcopy 
\def\aclpaperid{3366} 

\title{Tetra-Tagging: Word-Synchronous Parsing with Linear-Time Inference}

\author{Nikita Kitaev \and Dan Klein \\
  Computer Science Division \\
  University of California, Berkeley \\
  {\tt \{kitaev, klein\}@cs.berkeley.edu}}

\date{}

\begin{document}
\maketitle
\begin{abstract}
We  present  a  constituency  parsing  algorithm that, like a supertagger, works by assigning labels  to  each  word  in  a  sentence.   In order to maximally leverage current neural architectures, the model scores each word's tags in parallel, with minimal task-specific structure.  After scoring, a left-to-right reconciliation phase extracts a tree in (empirically) linear time.  Our parser achieves 95.4 F1 on the WSJ test set while also achieving substantial speedups compared to current state-of-the-art parsers with comparable accuracies.

\end{abstract}

\section{Introduction}

Recent progress in NLP, and practical machine learning applications more generally, has been driven in large part by increasing availability of compute. These advances are made possible by an ecosystem of specialized hardware accelerators such as GPUs and TPUs, highly tuned kernels for executing particular operations, and the ability to amortize computational costs across tasks through approaches such as pre-training and multi-task learning. This places particular demands for a model to be efficient: it must parallelize, it must maximally use standard subcomponents that have been heavily optimized, but at the same time it must adequately incorporate task-specific insights and inductive biases.

Against this backdrop, constituency parsing stands as a task where custom architectures are prevalent and parallel execution is limited. State-of-the-art approaches use custom architecture components, such as the tree-structured networks of RNNG~\citep{dyer-etal-2016-recurrent} or the per-span MLPs in chart parsers~\citep{stern-etal-2017-minimal,kitaev-etal-2019-multilingual}. Approaches to inference range from autoregressive generation, to cubic-time CKY, to A* search -- none of which are readily parallelizable. Our goal is to demonstrate a parsing algorithm that makes effective use of the latest hardware. The desiderata for our approach are \emph{(a)} to maximize parallelism, \emph{(b)}~to minimize task-specific architecture design, and \emph{(c)} to lose as little accuracy as possible compared to a state-of-the-art highly-specialized model. To do this, we propose an algorithm that reduces parsing to tagging, where all tags are predicted in parallel using a standard model architecture such as BERT~\citep{devlin-etal-2019-bert}. Tagging is followed by a minimal inference procedure that is fast enough to schedule on the CPU because it runs in linear time with low constant factors (subject to mild assumptions).

\section{Related Work}

\paragraph{Label-based parsing} A variety of approaches have been proposed to mostly or entirely reduce parsing to a sequence labeling task. One family of these approaches is \emph{supertagging}~\citep{bangalore-joshi-1999-supertagging}, which is particularly common for CCG parsing. CCG imposes constraints on which supertags may form a valid derivation, necessitating complex search procedures for finding a high-scoring sequence of supertags that is self-consistent. An example of how such a search procedure can be implemented is the system of \citet{lee-etal-2016-global}, which uses A$^*$ search.
This search procedure is not easily parallelizable on GPU-like hardware, and has a worst-case serial running time that is exponential in the sentence length. \citet{gomez-rodriguez-vilares-2018-constituent} propose a different approach that fully reduces parsing to sequence labeling, but the label set size is unbounded: it expands with tree depth and related properties of the input, rather than being fixed for any given language. There have been attempts to address this by adding redundant labels, where the model learns to switch between tagging schemes in an attempt to avoid the problem of unseen labels \citep{vilares-etal-2019-better}, but that only increases the label inventory rather than restricting it to a finite set. Our approach, on the other hand, uses just 4 labels in its simplest formulation (hence the name \emph{tetra-tagging}).

\paragraph{Shift-reduce transition systems} A number of parsers proposed in the literature can be categorized as \emph{shift-reduce} parsers~\citep{henderson-2003-inducing,sagae-lavie-2005-classifier,zhang-clark-2009-transition,zhu-etal-2013-fast}. These systems rely on generating sequences of actions, which need not be evenly distributed throughout the sentence. For example, the construction of a deep right-branching tree might involve a series of \emph{shift} actions (one per word in the sentence), followed by equally many consecutive \emph{reduce} actions that all cluster at the end of the sentence. Due to the uneven alignment between actions and locations in a sentence, neural network architectures in recent shift-reduce systems~\citep{vinyals-etal-2015-grammar,dyer-etal-2016-recurrent,liu-zhang-2017-order} generally follow an encoder-decoder approach with autoregressive generation rather than directly assigning labels to positions in the input. Our proposed parser is also transition-based, but there are guaranteed to be exactly two decisions to make between one word and the next. This fixed alignment allows us to predict all actions in parallel rather than autoregressively.

\paragraph{Chart parsing} Chart parsers fundamentally operate over \emph{span-aligned} rather than \emph{word-aligned} representations.
For instance, the size of the chart in the CKY algorithm~\citep{cocke-1970-programming,kasami-1966-efficient,younger-1967-recognition} is quadratic in the length of the sentence, and the algorithm itself has cubic running time.
This is true for both classical methods and more recent neural approaches~\citep{durrett-klein-2015-neural,stern-etal-2017-minimal}. The construction of a chart involves a non-trivial (quadratic) computation that is specialized to parsing, and implementing the CKY algorithm on a hardware accelerator is a nontrivial and hardware-specific task.

\paragraph{Left-corner parsing}
To achieve all of our desiderata, we combine aspects of the previously-mentioned approaches with ideas drawn from a long line of work on left-corner parsing~(\citealp{rosenkrantz-lewis-1970-deterministic,nijholt-1979-structure,vanschijndel-etal-2013-model,noji-etal-2016-using,shain-etal-2016-memory-bounded}, \emph{inter alia}). Much of past work highlights the benefits of a left-corner formulation for \emph{memory efficiency}, with implications for psycholinguistic plausibility of the approach. We, on the other hand, demonstrate how to leverage these same considerations to achieve parallel tagging and linear \emph{time complexity} of the subsequent inference procedure.
Further, past work has used grammars~\citep{rosenkrantz-lewis-1970-deterministic}, or transformed labeled trees~\citep{johnson-1998-finite-state,schuler-etal-2010-broad}. On the other hand, it is precisely the lack of an explicit grammar that allows us to formulate our linear-time inference algorithm.

\section{Method}

To introduce our method, we first restrict ourselves to only consider unlabeled full binary trees (where every node has either 0 or 2 children).
We defer the discussion of labeling and non-binary structure to Section~\ref{subsec:nary}.

\subsection{Trees to tags}

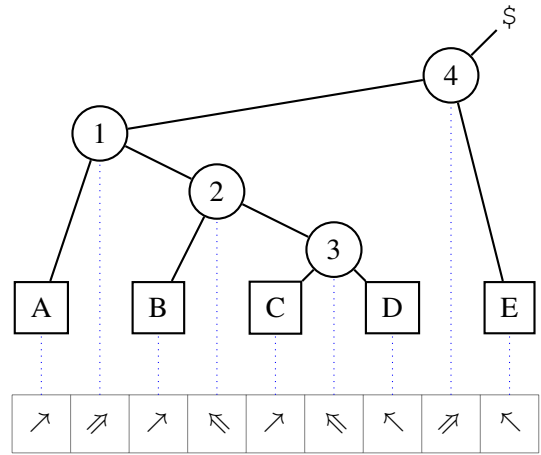
\begin{figure}
\center
\begin{tikzpicture}[x=0.1\linewidth, y=0.1\linewidth][%
\draw[] (1.5, 5.5) node[shape=circle,draw,thick,inner sep=4pt] (N1) {1};
\draw[] (3.5, 4.5) node[shape=circle,draw,thick,inner sep=4pt] (N2) {2};
\draw[] (5.5, 3.5) node[shape=circle,draw,thick,inner sep=4pt] (N3) {3};
\draw[] (7.5, 6.5) node[shape=circle,draw,thick,inner sep=4pt] (N4) {4};
\draw[] (8.5, 7.5) node[inner sep=1pt] (NR) {\large\tt \$};

\draw[] (0.5, 2.5) node[shape=rectangle,draw,thick,inner sep=6pt] (NA) {A};
\draw[] (2.5, 2.5) node[shape=rectangle,draw,thick,inner sep=6pt] (NB) {B};
\draw[] (4.5, 2.5) node[shape=rectangle,draw,thick,inner sep=6pt] (NC) {C};
\draw[] (6.5, 2.5) node[shape=rectangle,draw,thick,inner sep=6pt] (ND) {D};
\draw[] (8.5, 2.5) node[shape=rectangle,draw,thick,inner sep=6pt] (NE) {E};

\draw[thick] (NA) -- (N1);
\draw[thick] (NB) -- (N2);
\draw[thick] (NC) -- (N3);
\draw[thick] (ND) -- (N3);
\draw[thick] (NE) -- (N4);
\draw[thick] (N1) -- (N4);
\draw[thick] (N2) -- (N1);
\draw[thick] (N3) -- (N2);
\draw[thick] (N4) -- (NR);

\draw[step=1,gray,very thin] (0, 0) grid (9,1);
\draw[] (0.5, 0.5) node[] (LA) {\xl};
\draw[] (1.5, 0.5) node[] (L1) {\xL};
\draw[] (2.5, 0.5) node[] (LB) {\xl};
\draw[] (3.5, 0.5) node[] (L2) {\xR};
\draw[] (4.5, 0.5) node[] (LC) {\xl};
\draw[] (5.5, 0.5) node[] (L3) {\xR};
\draw[] (6.5, 0.5) node[] (LD) {\xr};
\draw[] (7.5, 0.5) node[] (L4) {\xL};
\draw[] (8.5, 0.5) node[] (LE) {\xr};

\draw[dotted,blue] (0.5, 1.0) -- (NA);
\draw[dotted,blue] (1.5, 1.0) -- (N1);
\draw[dotted,blue] (2.5, 1.0) -- (NB);
\draw[dotted,blue] (3.5, 1.0) -- (N2);
\draw[dotted,blue] (4.5, 1.0) -- (NC);
\draw[dotted,blue] (5.5, 1.0) -- (N3);
\draw[dotted,blue] (6.5, 1.0) -- (ND);
\draw[dotted,blue] (7.5, 1.0) -- (N4);
\draw[dotted,blue] (8.5, 1.0) -- (NE);

\end{tikzpicture}%

  \caption{\label{fig:scheme_overview} An example tree with the corresponding labels. The nonterminal nodes have been numbered based on an in-order traversal.}
\end{figure}

Consider the example tree shown in Figure~\ref{fig:scheme_overview}. The tree is fully binarized and consists of 5 terminal symbols (A,B,C,D,E) and 4 nonterminal nodes (1,2,3,4). For any full binary parse tree, the number of nonterminals will always be one less than the number of words, so we can construct a one-to-one mapping between nonterminals and fenceposts (i.e.\ positions between words): each fencepost is matched with the shortest span that crosses it.

For each node, we calculate the \emph{direction of its parent}, i.e.\ whether the node is a left-child or a right-child. Although the root node in the tree does not have a parent, by convention we treat it as though it were a left-child (in Figure~\ref{fig:scheme_overview}, this is denoted by the dummy parent labeled \texttt{\$}).

\pagebreak
Our scheme associates each word and fencepost in the sentence with one of four labels:
\begin{itemize}
\item ``\xl'': This terminal node is a left-child.
\item ``\xr'': This terminal node is a right-child.
\item ``\xL'': The shortest span crossing this fencepost is a left-child.
\item ``\xR'': The shortest span crossing this fencepost is a right-child.
\end{itemize}
We refer to our method as \textbf{tetra-tagging} because it uses only these four labels to represent binary bracketing structure.

\subsection{Model}

Given a sentence with $n$ words, there are altogether $2n-1$ decisions (each with two options). By the construction above, it is evident that every tree has one (and only one) corresponding label representation.
To reduce parsing to tagging, we simply use a neural network to predict which tag to select for each of the $2n-1$ decisions required.

Our implementation predicts these tag sequences from pre-trained BERT word representations. Two independent projection matrices are applied to the feature vector for the last sub-word unit within each word: one projection produces scores for actions corresponding to that word, and the other for actions at the following fencepost. A softmax loss is applied, and the model is trained to maximize the likelihood of the correct action sequence.

\subsection{Tags to trees: transition system}

To map from label sequences back to trees,
we re-interpret the four labels (``\xl'', ``\xr'', ``\xL'', ``\xR'') as actions in a left-corner transition system. The transition system maintains a \emph{stack} of partially-constructed trees, where each element of the stack is one of the following: (a) a terminal symbol, i.e.\ a word; (b) a complete tree; or (c) a tree with a single empty slot, denoted by the special element $\varnothing$. An empty slot must be the rightmost leaf node in its tree, but may occur at any depth.

\algnewcommand\algorithmicswitch{\textbf{switch}}
\algnewcommand\algorithmiccase{\textbf{case}}
\algdef{SE}[SWITCH]{Switch}{EndSwitch}[1]{\algorithmicswitch\ #1\ \algorithmicdo}{\algorithmicend\ \algorithmicswitch}%
\algdef{SE}[CASE]{Case}{EndCase}[1]{\algorithmiccase\ #1}{\algorithmicend\ \algorithmiccase}%

\begin{algorithm}[t]
\caption{Decoding algorithm}\label{algo:decode}
\renewcommand{\algorithmicrequire}{\textbf{Input:}}
\renewcommand{\algorithmicensure}{\textbf{Output:}}
\begin{algorithmic}[1]
\small
\Require{A list of words (\emph{words}) and a corresponding list of tetra-tags (\emph{actions})}
\Ensure{A parse tree}
    \State \emph{stack} $\gets$ []
    \State \emph{buffer} $\gets$ \emph{words}
    \For{\emph{action} in \emph{actions}}
        \Switch{\emph{action}}
            \Case{``\xl''}
              \State \emph{leaf} $\gets$ {\sc Pop-First}(\emph{buffer}) 
              \State \emph{stack} $\gets$ {\sc Push-Last}(\emph{stack}, \emph{leaf})
            \EndCase
            \Case{``\xr''}
              \State \emph{leaf} $\gets$ {\sc Pop-First}(\emph{buffer}) 
              \State \emph{stack}[$-1$] $\gets$ {\sc Combine}(\emph{stack}[$-1$], leaf) 
            \EndCase
            \Case{``\xL''}
              \State \emph{stack}[$-1$] $\gets$ {\sc Make-Node}(\emph{stack}[$-1$], $\varnothing$)
            \EndCase
            \Case{``\xR''}
              \State \emph{tree} $\gets$ {\sc Pop-Last}(\emph{stack})
              \State \emph{tree} $\gets$ {\sc Make-Node}(tree, $\varnothing$)
              \State \emph{stack}[$-1$] $\gets$ {\sc Combine}(\emph{stack}[$-1$], tree)
            \EndCase
          \EndSwitch
    \EndFor \Comment{The stack should only have one element}
    \State \textbf{return} \emph{stack}[$0$]
\end{algorithmic}
\end{algorithm}

The tree operations used are:
\emph{(a)}
{\sc Make-Node}(\emph{left-child}, \emph{right-child}), which creates a new tree node; and
\emph{(b)}
{\sc Combine}(\emph{parent-tree}, \emph{child-tree}), which replaces the empty slot $\varnothing$ in the parent tree with the child tree.

Decoding uses Algorithm~\ref{algo:decode}; an example derivation is shown in Figure~\ref{fig:sample_derivation}.

Each action in the transition system is responsible for adding a single tree node onto the stack: the actions ``\xl'' and ``\xr'' do this by shifting in a leaf node, while the actions ``\xL'' and ``\xR'' construct a new non-terminal node. The transition system maintains the invariant that the topmost stack element is a complete tree if and only if a leaf node was just shifted (i.e.\ the last action was either ``\xl'' or ``\xr''), and all other stack elements have a single empty slot.

The actions ``\xr'' and ``\xR'' both make use of the {\sc Combine} operation to fill an empty slot on the stack with a newly-introduced node, which makes the new node a right-child. New nodes from the actions ``\xl'' and ``\xL'', on the other hand, are introduced directly onto the stack and can become left-children via a later {\sc Make-Node} operation. As a result, the behavior of the four actions (``\xl'', ``\xr'', ``\xL'', ``\xR'') matches the label definitions from the previous section.

\subsection{Inference}
\label{subsec:inference}

The goal of inference is to select the sequence of labels that is assigned the highest probability by the tagging model. It should be noted that not all sequences of labels are valid under our transition system.
In particular:
\begin{itemize}
\item
The first action must be ``\xl'', because the stack is initially empty and the only valid action is to shift the first word in the sentence from the buffer onto the stack.
\item
The action ``\xR'' relies on there being more than one element on the stack (lines {17-19} of Algorithm~\ref{algo:decode}).
\item
After executing all actions, the stack should contain a single element. Due to the invariant that the top stack element after a ``\xl'' or ``\xr'' action is always a tree with no empty slots, this single stack element is guaranteed to be a complete tree that spans the full sentence.
\end{itemize}

We observe that the validity constraints for our transition system can be expressed entirely in terms of the \emph{number} of stack elements at each point in the derivation, and do not depend on the precise structure of those elements. This property enables an optimal and efficient dynamic program for finding the valid sequence of labels that has the highest probability under the model.

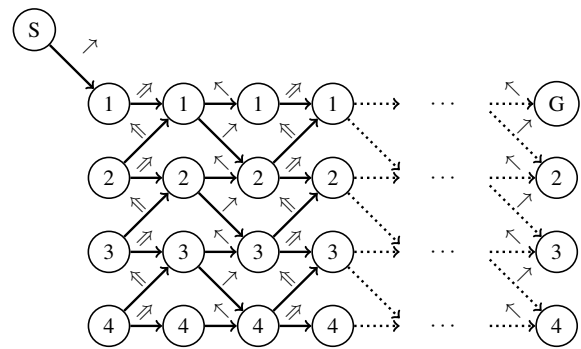
\begin{figure}[p]
\center
\scalebox{0.75}{
\begin{tikzpicture}[x=0.17\linewidth, y=0.17\linewidth][%
\draw[] (0.5, 5.5) node[shape=circle,draw,thick,inner sep=4pt] (A0) {S};

\draw[] (1.5, 4.5) node[shape=circle,draw,thick,inner sep=4pt] (B1) {1};
\draw[] (1.5, 3.5) node[shape=circle,draw,thick,inner sep=4pt] (B2) {2};
\draw[] (1.5, 2.5) node[shape=circle,draw,thick,inner sep=4pt] (B3) {3};
\draw[] (1.5, 1.5) node[shape=circle,draw,thick,inner sep=4pt] (B4) {4};

\draw[] (2.5, 4.5) node[shape=circle,draw,thick,inner sep=4pt] (C1) {1};
\draw[] (2.5, 3.5) node[shape=circle,draw,thick,inner sep=4pt] (C2) {2};
\draw[] (2.5, 2.5) node[shape=circle,draw,thick,inner sep=4pt] (C3) {3};
\draw[] (2.5, 1.5) node[shape=circle,draw,thick,inner sep=4pt] (C4) {4};

\draw[] (3.5, 4.5) node[shape=circle,draw,thick,inner sep=4pt] (D1) {1};
\draw[] (3.5, 3.5) node[shape=circle,draw,thick,inner sep=4pt] (D2) {2};
\draw[] (3.5, 2.5) node[shape=circle,draw,thick,inner sep=4pt] (D3) {3};
\draw[] (3.5, 1.5) node[shape=circle,draw,thick,inner sep=4pt] (D4) {4};

\draw[] (4.5, 4.5) node[shape=circle,draw,thick,inner sep=4pt] (E1) {1};
\draw[] (4.5, 3.5) node[shape=circle,draw,thick,inner sep=4pt] (E2) {2};
\draw[] (4.5, 2.5) node[shape=circle,draw,thick,inner sep=4pt] (E3) {3};
\draw[] (4.5, 1.5) node[shape=circle,draw,thick,inner sep=4pt] (E4) {4};

\draw[] (5.5, 4.5) node[] (F1) {};
\draw[] (5.5, 3.5) node[] (F2) {};
\draw[] (5.5, 2.5) node[] (F3) {};
\draw[] (5.5, 1.5) node[] (F4) {};

\draw[] (6, 4.5) node[] (DOT1) {$\cdots$};
\draw[] (6, 3.5) node[] (DOT2) {$\cdots$};
\draw[] (6, 2.5) node[] (DOT3) {$\cdots$};
\draw[] (6, 1.5) node[] (DOT4) {$\cdots$};

\draw[] (6.5, 4.5) node[] (G1) {};
\draw[] (6.5, 3.5) node[] (G2) {};
\draw[] (6.5, 2.5) node[] (G3) {};
\draw[] (6.5, 1.5) node[] (G4) {};

\draw[] (7.5, 4.5) node[shape=circle,draw,thick,inner sep=4pt] (H1) {G};
\draw[] (7.5, 3.5) node[shape=circle,draw,thick,inner sep=4pt] (H2) {2};
\draw[] (7.5, 2.5) node[shape=circle,draw,thick,inner sep=4pt] (H3) {3};
\draw[] (7.5, 1.5) node[shape=circle,draw,thick,inner sep=4pt] (H4) {4};

\draw[very thick,->] (A0) -- node[anchor=south west]{\xl} (B1);

\draw[very thick,->] (B1) -- node[midway,above,inner sep=1pt]{\xL} (C1);
\draw[very thick,->] (B2) -- node[midway,above,inner sep=1pt]{\xL} (C2);
\draw[very thick,->] (B3) -- node[midway,above,inner sep=1pt]{\xL} (C3);
\draw[very thick,->] (B4) -- node[midway,above,inner sep=1pt]{\xL} (C4);

\draw[very thick,->] (B2) -- node[anchor=south east,inner sep=-1pt]{\xR} (C1);
\draw[very thick,->] (B3) -- node[anchor=south east,inner sep=-1pt]{\xR} (C2);
\draw[very thick,->] (B4) -- node[anchor=south east,inner sep=-1pt]{\xR} (C3);

\draw[very thick,->] (C1) -- node[midway,above,inner sep=1pt]{\xr} (D1);
\draw[very thick,->] (C2) -- node[midway,above,inner sep=1pt]{\xr} (D2);
\draw[very thick,->] (C3) -- node[midway,above,inner sep=1pt]{\xr} (D3);
\draw[very thick,->] (C4) -- node[midway,above,inner sep=1pt]{\xr} (D4);

\draw[very thick,->] (C1) -- node[anchor=south west,inner sep=-1pt]{\xl} (D2);
\draw[very thick,->] (C2) -- node[anchor=south west,inner sep=-1pt]{\xl} (D3);
\draw[very thick,->] (C3) -- node[anchor=south west,inner sep=-1pt]{\xl} (D4);

\draw[very thick,->] (D1) -- node[midway,above,inner sep=1pt]{\xL} (E1);
\draw[very thick,->] (D2) -- node[midway,above,inner sep=1pt]{\xL} (E2);
\draw[very thick,->] (D3) -- node[midway,above,inner sep=1pt]{\xL} (E3);
\draw[very thick,->] (D4) -- node[midway,above,inner sep=1pt]{\xL} (E4);

\draw[very thick,->] (D2) -- node[anchor=south east,inner sep=-1pt]{\xR} (E1);
\draw[very thick,->] (D3) -- node[anchor=south east,inner sep=-1pt]{\xR} (E2);
\draw[very thick,->] (D4) -- node[anchor=south east,inner sep=-1pt]{\xR} (E3);

\draw[very thick,dotted,->] (E1) -- (F1);
\draw[very thick,dotted,->] (E2) -- (F2);
\draw[very thick,dotted,->] (E3) -- (F3);
\draw[very thick,dotted,->] (E4) -- (F4);

\draw[very thick,dotted,->] (E1) -- (F2);
\draw[very thick,dotted,->] (E2) -- (F3);
\draw[very thick,dotted,->] (E3) -- (F4);

\draw[very thick,dotted,->] (G1) -- node[midway,above,inner sep=1pt]{\xr} (H1);
\draw[very thick,dotted,->] (G2) -- node[midway,above,inner sep=1pt]{\xr} (H2);
\draw[very thick,dotted,->] (G3) -- node[midway,above,inner sep=1pt]{\xr} (H3);
\draw[very thick,dotted,->] (G4) -- node[midway,above,inner sep=1pt]{\xr} (H4);

\draw[very thick,dotted,->] (G1) -- node[anchor=south west,inner sep=-1pt]{\xl} (H2);
\draw[very thick,dotted,->] (G2) -- node[anchor=south west,inner sep=-1pt]{\xl} (H3);
\draw[very thick,dotted,->] (G3) -- node[anchor=south west,inner sep=-1pt]{\xl} (H4);

\end{tikzpicture}}
 \caption{\label{fig:inference} Paths in this grid correspond to sequences of tags, where paths starting at S and arriving at G are valid trees. Numbers represent the number of elements on the stack.}
\end{figure}

The dynamic program maintains a table of the highest-scoring parser state for each combination of \emph{number of actions taken} and \emph{stack depth}. Prior to taking any actions, the stack must be empty. The algorithm then proceeds left-to-right through the sentence to fill in highest-scoring stack configurations after action 1, 2, etc. The dynamic program can be visualized as finding the shortest path through a graph like Figure~\ref{fig:inference}, where each action-count/stack-depth combination is represented by a node, and a transition is represented by an edge with weight equal to the model-predicted score of the associated tag.

The time complexity of this dynamic program depends on the number of actions (which is $2n-1$, where $n$ is the length of the sentence), as well as the maximum possible depth of the stack ($d$). A left-corner transition system has the property that stack depth tends to be small for parse trees of natural language~\citep{abney-johnson-1991-memory,schuler-etal-2010-broad}. In practice, the largest stack depth observed at any point in the derivation for any tree in the Penn Treebank is 8. By comparison, the median sentence length in the data is 23, and the longest sentence contains over 100 words.

As a result, we can cap the maximum stack depth allowed in our inference procedure to $d = 8$, which means that the $O(nd^2)$ time complexity of inference is effectively $O(n)$. In other words, our inference procedure will, in practice, take linear time in the length of the sentence.


\newcommand{\newstacka}[0]{%
\small
\begin{tikzpicture}[x=0.05\linewidth, y=0.05\linewidth][%
\draw[] (1.5, 1.5) node[shape=circle,thick,inner sep=0pt,blue] (N1) {\tt \$};

\draw[] (0.3, 0.5) node[shape=rectangle,blue,draw,thick,inner sep=2pt] (NA) {A};

\draw[thick,blue] (NA) -- (N1);
\end{tikzpicture}}

\newcommand{\newstackb}[0]{%
\small
\begin{tikzpicture}[x=0.05\linewidth, y=0.05\linewidth][%
\draw[] (1.5, 1.5) node[shape=circle,thick,inner sep=0pt,blue] (N1) {\tt \$};

\draw[] (0.3, 0.5) node[shape=rectangle,blue,draw,thick,inner sep=2pt] (NA) {B};

\draw[thick,blue] (NA) -- (N1);
\end{tikzpicture}}

\newcommand{\newstackc}[0]{%
\small
\begin{tikzpicture}[x=0.05\linewidth, y=0.05\linewidth][%
\draw[] (1.5, 1.5) node[shape=circle,thick,inner sep=0pt,blue] (N1) {\tt \$};

\draw[] (0.3, 0.5) node[shape=rectangle,blue,draw,thick,inner sep=2pt] (NA) {C};

\draw[thick,blue] (NA) -- (N1);
\end{tikzpicture}}

\newcommand{\newstackaa}[0]{%
\small
\begin{tikzpicture}[x=0.05\linewidth, y=0.05\linewidth][%
\draw[] (1.5, 1.5) node[shape=circle,blue,draw,thick,inner sep=1pt] (N1) {1};
\draw[] (2.7, 0.3) node[blue,inner sep=1pt] (N2) {$\varnothing$};

\draw[] (2.5, 2.5) node[blue,inner sep=0pt] (N4) {\tt \$};

\draw[] (0.5, 0.3) node[shape=rectangle,draw,thick,inner sep=2pt] (NA) {A};

\draw[thick] (NA) -- (N1);
\draw[thick,blue] (N1) -- (N4);
\draw[thick,blue] (N2) -- (N1);
\end{tikzpicture}}

\newcommand{\stackaa}[0]{%
\small
\begin{tikzpicture}[x=0.05\linewidth, y=0.05\linewidth][%
\draw[] (1.5, 1.5) node[shape=circle,draw,thick,inner sep=1pt] (N1) {1};
\draw[] (2.7, 0.3) node[inner sep=1pt] (N2) {$\varnothing$};

\draw[] (2.5, 2.5) node[inner sep=0pt] (N4) {\tt \$};

\draw[] (0.5, 0.3) node[shape=rectangle,draw,thick,inner sep=2pt] (NA) {A};

\draw[thick] (NA) -- (N1);
\draw[thick] (N1) -- (N4);
\draw[thick] (N2) -- (N1);
\end{tikzpicture}}


\newcommand{\newstackaabb}[0]{%
\small
\begin{tikzpicture}[x=0.05\linewidth, y=0.05\linewidth][%
\draw[] (1.5, 2.5) node[shape=circle,draw,thick,inner sep=1pt] (N1) {1};
\draw[] (3.5, 1.5) node[shape=circle,blue,draw,thick,inner sep=1pt] (N2) {2};
\draw[] (4.7, 0.3) node[blue,inner sep=1pt] (N3) {$\varnothing$};
\draw[] (4.5, 3.5) node[inner sep=0pt] (N4) {\tt \$};

\draw[] (0.5, 0.3) node[shape=rectangle,draw,thick,inner sep=2pt] (NA) {A};
\draw[] (2.5, 0.3) node[shape=rectangle,draw,thick,inner sep=2pt] (NB) {B};

\draw[thick] (NA) -- (N1);
\draw[thick] (NB) -- (N2);
\draw[thick] (N1) -- (N4);
\draw[thick,blue] (N2) -- (N1);
\draw[thick,blue] (N3) -- (N2);
\end{tikzpicture}}

\newcommand{\stackaabb}[0]{%
\small
\begin{tikzpicture}[x=0.05\linewidth, y=0.05\linewidth][%
\draw[] (1.5, 2.5) node[shape=circle,draw,thick,inner sep=1pt] (N1) {1};
\draw[] (3.5, 1.5) node[shape=circle,draw,thick,inner sep=1pt] (N2) {2};
\draw[] (4.7, 0.3) node[inner sep=1pt] (N3) {$\varnothing$};
\draw[] (4.5, 3.5) node[inner sep=0pt] (N4) {\tt \$};

\draw[] (0.5, 0.3) node[shape=rectangle,draw,thick,inner sep=2pt] (NA) {A};
\draw[] (2.5, 0.3) node[shape=rectangle,draw,thick,inner sep=2pt] (NB) {B};

\draw[thick] (NA) -- (N1);
\draw[thick] (NB) -- (N2);
\draw[thick] (N1) -- (N4);
\draw[thick] (N2) -- (N1);
\draw[thick] (N3) -- (N2);
\end{tikzpicture}}


\newcommand{\newstackaabbcc}[0]{%
\small
\begin{tikzpicture}[x=0.05\linewidth, y=0.05\linewidth][%
\draw[] (1.5, 3.5) node[shape=circle,draw,thick,inner sep=1pt] (N1) {1};
\draw[] (3.5, 2.5) node[shape=circle,draw,thick,inner sep=1pt] (N2) {2};
\draw[] (5.5, 1.5) node[shape=circle,draw,blue,thick,inner sep=1pt] (N3) {3};
\draw[] (6.5, 4.5) node[inner sep=0pt] (N4) {\tt \$};

\draw[] (0.5, 0.3) node[shape=rectangle,draw,thick,inner sep=2pt] (NA) {A};
\draw[] (2.5, 0.3) node[shape=rectangle,draw,thick,inner sep=2pt] (NB) {B};
\draw[] (4.5, 0.3) node[shape=rectangle,draw,thick,inner sep=2pt] (NC) {C};
\draw[] (6.5, 0.3) node[blue,inner sep=1pt] (ND) {$\varnothing$};

\draw[thick] (NA) -- (N1);
\draw[thick] (NB) -- (N2);
\draw[thick] (NC) -- (N3);
\draw[thick,blue] (ND) -- (N3);
\draw[thick] (N1) -- (N4);
\draw[thick] (N2) -- (N1);
\draw[thick,blue] (N3) -- (N2);
\end{tikzpicture}}

\newcommand{\newstackaabbccd}[0]{%
\small
\begin{tikzpicture}[x=0.05\linewidth, y=0.05\linewidth][%
\draw[] (1.5, 3.5) node[shape=circle,draw,thick,inner sep=1pt] (N1) {1};
\draw[] (3.5, 2.5) node[shape=circle,draw,thick,inner sep=1pt] (N2) {2};
\draw[] (5.5, 1.5) node[shape=circle,draw,thick,inner sep=1pt] (N3) {3};
\draw[] (7.5, 4.5) node[inner sep=0pt] (N4) {\tt \$};

\draw[] (0.5, 0.3) node[shape=rectangle,draw,thick,inner sep=2pt] (NA) {A};
\draw[] (2.5, 0.3) node[shape=rectangle,draw,thick,inner sep=2pt] (NB) {B};
\draw[] (4.5, 0.3) node[shape=rectangle,draw,thick,inner sep=2pt] (NC) {C};
\draw[] (6.5, 0.3) node[shape=rectangle,draw,blue,thick,inner sep=2pt] (ND) {D};

\draw[thick] (NA) -- (N1);
\draw[thick] (NB) -- (N2);
\draw[thick] (NC) -- (N3);
\draw[thick,blue] (ND) -- (N3);
\draw[thick] (N1) -- (N4);
\draw[thick] (N2) -- (N1);
\draw[thick] (N3) -- (N2);
\end{tikzpicture}}


\newcommand{\newstackaabbccdd}[0]{%
\small
\begin{tikzpicture}[x=0.05\linewidth, y=0.05\linewidth][%
\draw[] (1.5, 3.5) node[shape=circle,draw,thick,inner sep=1pt] (N1) {1};
\draw[] (3.5, 2.5) node[shape=circle,draw,thick,inner sep=1pt] (N2) {2};
\draw[] (5.5, 1.5) node[shape=circle,draw,thick,inner sep=1pt] (N3) {3};
\draw[] (7.5, 4.5) node[shape=circle,draw,blue,thick,inner sep=1pt] (N4) {4};
\draw[] (8.5, 5.5) node[blue,inner sep=0pt] (NR) {\tt \$};

\draw[] (0.5, 0.3) node[shape=rectangle,draw,thick,inner sep=2pt] (NA) {A};
\draw[] (2.5, 0.3) node[shape=rectangle,draw,thick,inner sep=2pt] (NB) {B};
\draw[] (4.5, 0.3) node[shape=rectangle,draw,thick,inner sep=2pt] (NC) {C};
\draw[] (6.5, 0.3) node[shape=rectangle,draw,thick,inner sep=2pt] (ND) {D};
\draw[] (8.5, 3.3) node[blue,inner sep=1pt] (NE) {$\varnothing$};

\draw[thick] (NA) -- (N1);
\draw[thick] (NB) -- (N2);
\draw[thick] (NC) -- (N3);
\draw[thick] (ND) -- (N3);
\draw[thick,blue] (NE) -- (N4);
\draw[thick] (N1) -- (N4);
\draw[thick] (N2) -- (N1);
\draw[thick] (N3) -- (N2);
\draw[thick,blue] (N4) -- (NR);
\end{tikzpicture}}

\newcommand{\newstackaabbccdde}[0]{%
\small
\begin{tikzpicture}[x=0.05\linewidth, y=0.05\linewidth][%
\draw[] (1.5, 3.5) node[shape=circle,draw,thick,inner sep=1pt] (N1) {1};
\draw[] (3.5, 2.5) node[shape=circle,draw,thick,inner sep=1pt] (N2) {2};
\draw[] (5.5, 1.5) node[shape=circle,draw,thick,inner sep=1pt] (N3) {3};
\draw[] (7.5, 4.5) node[shape=circle,draw,thick,inner sep=1pt] (N4) {4};
\draw[] (8.5, 5.5) node[inner sep=0pt] (NR) {\tt \$};

\draw[] (0.5, 0.3) node[shape=rectangle,draw,thick,inner sep=2pt] (NA) {A};
\draw[] (2.5, 0.3) node[shape=rectangle,draw,thick,inner sep=2pt] (NB) {B};
\draw[] (4.5, 0.3) node[shape=rectangle,draw,thick,inner sep=2pt] (NC) {C};
\draw[] (6.5, 0.3) node[shape=rectangle,draw,thick,inner sep=2pt] (ND) {D};
\draw[] (8.5, 0.3) node[shape=rectangle,draw,blue,thick,inner sep=2pt] (NE) {E};

\draw[thick] (NA) -- (N1);
\draw[thick] (NB) -- (N2);
\draw[thick] (NC) -- (N3);
\draw[thick] (ND) -- (N3);
\draw[thick,blue] (NE) -- (N4);
\draw[thick] (N1) -- (N4);
\draw[thick] (N2) -- (N1);
\draw[thick] (N3) -- (N2);
\draw[thick] (N4) -- (NR);
\end{tikzpicture}}
\newcommand{\dashrule}[1][black]{%
  \color{#1}\rule[\dimexpr.5ex-.2pt]{4pt}{.4pt}\xleaders\hbox{\rule{4pt}{0pt}\rule[\dimexpr.5ex-.2pt]{4pt}{.4pt}}\hfill\kern0pt%
}
\newcommand{\mydashrule}[0]{%
\\[-2\jot]
\multicolumn{3}{@{}c@{}}{\makebox[\linewidth]{\dashrule[black!60]}} \\[-\jot]}
\begin{figure}[p]
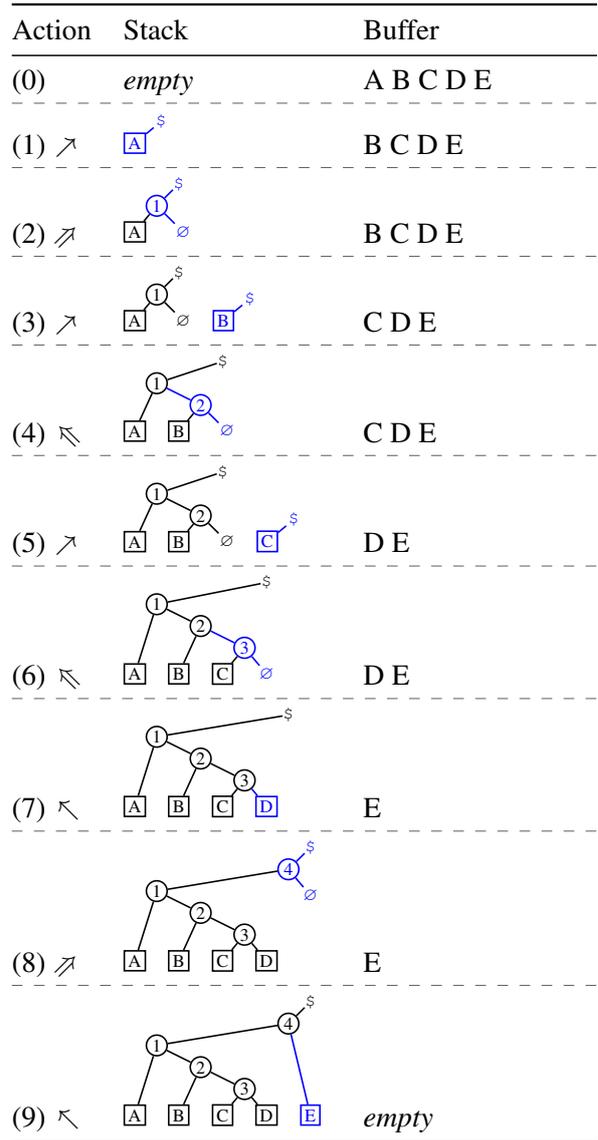

  \begin{center}
  \begin{tabular}{@{}l@{\hskip 12pt}l@{\hskip 16pt}l@{}}
    \toprule
    Action & Stack & Buffer  \\
    \midrule
    (0)     & \emph{empty} & A B C D E
    \mydashrule
    (1) \xl & \scalebox{0.75}{\newstacka} & B C D E
    \mydashrule
    (2) \xL & \scalebox{0.75}{\newstackaa} & B C D E
    \mydashrule
    (3) \xl & \scalebox{0.75}{\stackaa {\hskip 8pt} \newstackb} & C D E 
    \mydashrule
    (4) \xR & \scalebox{0.75}{\newstackaabb} & C D E 
    \mydashrule
    (5) \xl & \scalebox{0.75}{\stackaabb {\hskip 8pt} \newstackc} & D E 
    \mydashrule
    (6) \xR & \scalebox{0.75}{\newstackaabbcc} & D E 
    \mydashrule
    (7) \xr & \scalebox{0.75}{\newstackaabbccd} & E 
    \mydashrule
    (8) \xL & \scalebox{0.75}{\newstackaabbccdd} & E 
    \mydashrule
    (9) \xr & \scalebox{0.75}{\newstackaabbccdde} & \emph{empty}\\
    \bottomrule
  \end{tabular}
  \end{center}
  \caption{\label{fig:sample_derivation} An example derivation under our transition system.}
\end{figure}

\begin{table}[t]
\begin{center}
\resizebox{\linewidth}{!}{
\begin{tabular}{@{}lllc@{}}
\toprule
 & Sents/s & Hardware & F1  \\
\midrule
\citet{vilares-etal-2019-better} & \phantom{0}942 & 1x GPU & 91.13 \\
\citet{kitaev-etal-2019-multilingual}$^*$ & \phantom{00}39 & 1x GPU & 95.59 \\
\citet{zhou-zhao-2019-head}$^*$ & \multicolumn{1}{c}{--} & \multicolumn{1}{c}{--} & 95.84 \\
This work$^*$ & 1200 & 1x TPU v3-8 & 95.44 \\
\bottomrule
\end{tabular}
}
\end{center}
\caption{\label{table:results} Comparison of F1 scores and inference speeds on the WSJ test set. $^*$Models using BERT$_\texttt{LARGE}$ \citep{devlin-etal-2019-bert} word representations fine-tuned from the same initial parameters.}
\end{table}

\begin{figure}[t]
\centering
\includegraphics{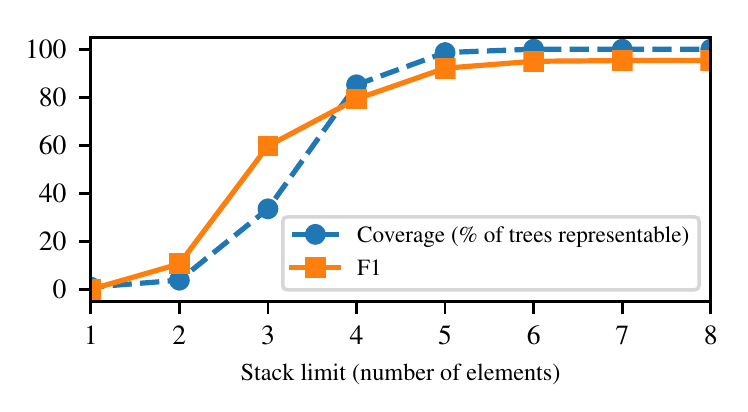}
\vspace{-9pt}
\caption{\label{fig:coverage} With a modest maximum stack size, the tetra-tagging transition system has near-complete coverage of the development data. Our parser's F1 score closely tracks the fraction of gold trees that can be represented.}
\end{figure}

\subsection{Handling of labels and non-binary trees}
\label{subsec:nary}

Each of our four actions creates a single node in the binary tree. Labeling a node can therefore be incorporated into the corresponding action; for example, the action ``\xL\texttt{S}'' will construct an \texttt{S} node that is a left-child in the tree.
We do not impose any constraints on valid label configurations, so our inference procedure remains virtually unchanged.

To handle non-binary trees, we first collapse all unary chains by introducing additional labels. For example, a clause that consists only of a verb phrase would be assigned the label \texttt{S-VP}. We then ensure that each non-terminal node has exactly two children by applying fully right-branching binarization, where a dummy label is introduced and assigned to nodes generated as a result of binarization. During inference, a post-processing step undoes these transformations.

\section{Results}

Our proposed parser is designed to rank syntactic decisions entirely in parallel, with inference reduced to a minimal linear-time algorithm. Its neural architecture consists almost entirely of BERT layers, with the only additions being two trainable projection matrices. To verify our approach, we train our parser on the Penn Treebank \citep{marcus-etal-1993-building} and evaluate its efficiency and accuracy when running on Cloud TPU v3 hardware.

In Table~\ref{table:results}, we compare with two classes of recent work. The parser by \citet{vilares-etal-2019-better} is one of the fastest reported in the recent literature, but it trails the state-of-the-art model by more than 4 F1 points. In contrast, models by \citet{zhou-zhao-2019-head} and \citet{kitaev-etal-2019-multilingual} achieve the highest-reported numbers when fine-tuning from the same initial BERT$_\texttt{LARGE}$ checkpoint that we use to train our tetra-tagger. However, these latter models are slower than our tetra-tagging approach and feature inference algorithms with high polynomial complexity that are difficult to adapt to accelerators such as the TPU.
Our approach is able to achieve both high throughput and high F1, with only small losses in accuracy compared to the best BERT-based approaches.

In Figure~\ref{fig:coverage}, we plot the parser's accuracy across different settings for the maximum stack depth. The F1 score rapidly asymptotes as the stack size limit is increased, which validates our claim that inference can run in linear time.

\section{Conclusion}

We present a reduction from constituency parsing to a tagging task with two binary structural decisions and two labeling decisions per word. Remarkably, probabilities for these tags can be estimated fully in parallel by a simple classification layer on top of a neural network architecture such as BERT.
We hope that this formulation can be useful as a simple and low-overhead way of integrating syntax into any neural NLP model, including for multi-task training and to predict syntactic annotations during inference. By reducing the task-specific architecture components to a minimum, our method can be rapidly adapted as new modeling techniques, efficiency optimizations, and hardware accelerators become available. Code for our approach is available at \href{https://github.com/nikitakit/tetra-tagging}{\fontsize{10}{\baselineskip}\selectfont\tt github.com/nikitakit/tetra-tagging}.

\section*{Acknowledgments}
This research was supported by DARPA through the XAI program and by the National Science Foundation under Grant No.\ 1618460. We would like to thank the Google Cloud TPU team for their hardware support. We are also grateful to the members of the Berkeley NLP group and the anonymous reviewers for their helpful feedback.

\bibliography{acl2020,anthology}

\begin{thebibliography}{28}
\expandafter\ifx\csname natexlab\endcsname\relax\def\natexlab#1{#1}\fi

\bibitem[{Abney and Johnson(1991)}]{abney-johnson-1991-memory}
Steven~P. Abney and Mark Johnson. 1991.
\newblock Memory requirements and local ambiguities of parsing strategies.
\newblock \emph{Journal of Psycholinguistic Research}, 20(3):233--250.

\bibitem[{Bangalore and Joshi(1999)}]{bangalore-joshi-1999-supertagging}
Srinivas Bangalore and Aravind~K. Joshi. 1999.
\newblock \href {https://www.aclweb.org/anthology/J99-2004} {{S}upertagging: An
  approach to almost parsing}.
\newblock \emph{Computational Linguistics}, 25(2):237--265.

\bibitem[{Cocke(1970)}]{cocke-1970-programming}
John Cocke. 1970.
\newblock Programming languages and their compilers: {{Preliminary}} notes.

\bibitem[{Devlin et~al.(2019)Devlin, Chang, Lee, and
  Toutanova}]{devlin-etal-2019-bert}
Jacob Devlin, Ming-Wei Chang, Kenton Lee, and Kristina Toutanova. 2019.
\newblock \href {https://doi.org/10.18653/v1/N19-1423} {{BERT}: Pre-training of
  deep bidirectional transformers for language understanding}.
\newblock In \emph{Proceedings of the 2019 Conference of the North {A}merican
  Chapter of the Association for Computational Linguistics: Human Language
  Technologies, Volume 1 (Long and Short Papers)}, pages 4171--4186,
  Minneapolis, Minnesota. Association for Computational Linguistics.

\bibitem[{Durrett and Klein(2015)}]{durrett-klein-2015-neural}
Greg Durrett and Dan Klein. 2015.
\newblock \href {https://doi.org/10.3115/v1/P15-1030} {Neural {CRF} parsing}.
\newblock In \emph{Proceedings of the 53rd Annual Meeting of the Association
  for Computational Linguistics and the 7th International Joint Conference on
  Natural Language Processing (Volume 1: Long Papers)}, pages 302--312,
  Beijing, China. Association for Computational Linguistics.

\bibitem[{Dyer et~al.(2016)Dyer, Kuncoro, Ballesteros, and
  Smith}]{dyer-etal-2016-recurrent}
Chris Dyer, Adhiguna Kuncoro, Miguel Ballesteros, and Noah~A. Smith. 2016.
\newblock \href {https://doi.org/10.18653/v1/N16-1024} {Recurrent neural
  network grammars}.
\newblock In \emph{Proceedings of the 2016 Conference of the North {A}merican
  Chapter of the Association for Computational Linguistics: Human Language
  Technologies}, pages 199--209, San Diego, California. Association for
  Computational Linguistics.

\bibitem[{G{\'o}mez-Rodr{\'\i}guez and
  Vilares(2018)}]{gomez-rodriguez-vilares-2018-constituent}
Carlos G{\'o}mez-Rodr{\'\i}guez and David Vilares. 2018.
\newblock \href {https://doi.org/10.18653/v1/D18-1162} {Constituent parsing as
  sequence labeling}.
\newblock In \emph{Proceedings of the 2018 Conference on Empirical Methods in
  Natural Language Processing}, pages 1314--1324, Brussels, Belgium.
  Association for Computational Linguistics.

\bibitem[{Henderson(2003)}]{henderson-2003-inducing}
James Henderson. 2003.
\newblock \href {https://www.aclweb.org/anthology/N03-1014} {Inducing history
  representations for broad coverage statistical parsing}.
\newblock In \emph{Proceedings of the 2003 Human Language Technology Conference
  of the North {A}merican Chapter of the Association for Computational
  Linguistics}, pages 103--110.

\bibitem[{Johnson(1998)}]{johnson-1998-finite-state}
Mark Johnson. 1998.
\newblock \href {https://www.aclweb.org/anthology/C98-1098} {Finite-state
  approximation of constraint-based grammars using left-corner grammar
  transforms}.
\newblock In \emph{{COLING} 1998 Volume 1: The 17th International Conference on
  Computational Linguistics}.

\bibitem[{Kasami(1966)}]{kasami-1966-efficient}
Tadao Kasami. 1966.
\newblock An efficient recognition and syntax-analysis algorithm for
  context-free languages.
\newblock \emph{Coordinated Science Laboratory Report no. R-257}.

\bibitem[{Kitaev et~al.(2019)Kitaev, Cao, and
  Klein}]{kitaev-etal-2019-multilingual}
Nikita Kitaev, Steven Cao, and Dan Klein. 2019.
\newblock \href {https://doi.org/10.18653/v1/P19-1340} {Multilingual
  constituency parsing with self-attention and pre-training}.
\newblock In \emph{Proceedings of the 57th Annual Meeting of the Association
  for Computational Linguistics}, pages 3499--3505, Florence, Italy.
  Association for Computational Linguistics.

\bibitem[{Lee et~al.(2016)Lee, Lewis, and Zettlemoyer}]{lee-etal-2016-global}
Kenton Lee, Mike Lewis, and Luke Zettlemoyer. 2016.
\newblock \href {https://doi.org/10.18653/v1/D16-1262} {Global neural {CCG}
  parsing with optimality guarantees}.
\newblock In \emph{Proceedings of the 2016 Conference on Empirical Methods in
  Natural Language Processing}, pages 2366--2376, Austin, Texas. Association
  for Computational Linguistics.

\bibitem[{Liu and Zhang(2017)}]{liu-zhang-2017-order}
Jiangming Liu and Yue Zhang. 2017.
\newblock \href {https://doi.org/10.1162/tacl_a_00070} {In-order
  transition-based constituent parsing}.
\newblock \emph{Transactions of the Association for Computational Linguistics},
  5:413--424.

\bibitem[{Marcus et~al.(1993)Marcus, Santorini, and
  Marcinkiewicz}]{marcus-etal-1993-building}
Mitchell~P. Marcus, Beatrice Santorini, and Mary~Ann Marcinkiewicz. 1993.
\newblock \href {https://www.aclweb.org/anthology/J93-2004} {Building a large
  annotated corpus of {E}nglish: The {P}enn {T}reebank}.
\newblock \emph{Computational Linguistics}, 19(2):313--330.

\bibitem[{Nijholt(1979)}]{nijholt-1979-structure}
Anton Nijholt. 1979.
\newblock Structure preserving transformations on non-left-recursive grammars.
\newblock In \emph{International {{Colloquium}} on {{Automata}}, {{Languages}},
  and {{Programming}}}, pages 446--459. {Springer}.

\bibitem[{Noji et~al.(2016)Noji, Miyao, and Johnson}]{noji-etal-2016-using}
Hiroshi Noji, Yusuke Miyao, and Mark Johnson. 2016.
\newblock \href {https://doi.org/10.18653/v1/D16-1004} {Using left-corner
  parsing to encode universal structural constraints in grammar induction}.
\newblock In \emph{Proceedings of the 2016 Conference on Empirical Methods in
  Natural Language Processing}, pages 33--43, Austin, Texas. Association for
  Computational Linguistics.

\bibitem[{Rosenkrantz and Lewis(1970)}]{rosenkrantz-lewis-1970-deterministic}
Daniel~J. Rosenkrantz and Philip~M. Lewis. 1970.
\newblock Deterministic left corner parsing.
\newblock In \emph{Switching and {{Automata Theory}}, 1970., {{IEEE Conference
  Record}} of 11th {{Annual Symposium}} On}, pages 139--152. {IEEE}.

\bibitem[{Sagae and Lavie(2005)}]{sagae-lavie-2005-classifier}
Kenji Sagae and Alon Lavie. 2005.
\newblock \href {https://www.aclweb.org/anthology/W05-1513} {A classifier-based
  parser with linear run-time complexity}.
\newblock In \emph{Proceedings of the Ninth International Workshop on Parsing
  Technology}, pages 125--132, Vancouver, British Columbia. Association for
  Computational Linguistics.

\bibitem[{Schuler et~al.(2010)Schuler, AbdelRahman, Miller, and
  Schwartz}]{schuler-etal-2010-broad}
William Schuler, Samir AbdelRahman, Tim Miller, and Lane Schwartz. 2010.
\newblock \href {https://doi.org/10.1162/coli.2010.36.1.36100} {Broad-coverage
  parsing using human-like memory constraints}.
\newblock \emph{Computational Linguistics}, 36(1):1--30.

\bibitem[{Shain et~al.(2016)Shain, Bryce, Jin, Krakovna, Doshi-Velez, Miller,
  Schuler, and Schwartz}]{shain-etal-2016-memory-bounded}
Cory Shain, William Bryce, Lifeng Jin, Victoria Krakovna, Finale Doshi-Velez,
  Timothy Miller, William Schuler, and Lane Schwartz. 2016.
\newblock \href {https://www.aclweb.org/anthology/C16-1092} {Memory-bounded
  left-corner unsupervised grammar induction on child-directed input}.
\newblock In \emph{Proceedings of {COLING} 2016, the 26th International
  Conference on Computational Linguistics: Technical Papers}, pages 964--975,
  Osaka, Japan. The COLING 2016 Organizing Committee.

\bibitem[{Stern et~al.(2017)Stern, Andreas, and
  Klein}]{stern-etal-2017-minimal}
Mitchell Stern, Jacob Andreas, and Dan Klein. 2017.
\newblock \href {https://doi.org/10.18653/v1/P17-1076} {A minimal span-based
  neural constituency parser}.
\newblock In \emph{Proceedings of the 55th Annual Meeting of the Association
  for Computational Linguistics (Volume 1: Long Papers)}, pages 818--827,
  Vancouver, Canada. Association for Computational Linguistics.

\bibitem[{{van Schijndel} et~al.(2013){van Schijndel}, Exley, and
  Schuler}]{vanschijndel-etal-2013-model}
Marten {van Schijndel}, Andy Exley, and William Schuler. 2013.
\newblock A model of language processing as hierarchic sequential prediction.
\newblock \emph{Topics in Cognitive Science}, 5(3):522--540.

\bibitem[{Vilares et~al.(2019)Vilares, Abdou, and
  S{\o}gaard}]{vilares-etal-2019-better}
David Vilares, Mostafa Abdou, and Anders S{\o}gaard. 2019.
\newblock \href {https://doi.org/10.18653/v1/N19-1341} {Better, faster,
  stronger sequence tagging constituent parsers}.
\newblock In \emph{Proceedings of the 2019 Conference of the North {A}merican
  Chapter of the Association for Computational Linguistics: Human Language
  Technologies, Volume 1 (Long and Short Papers)}, pages 3372--3383,
  Minneapolis, Minnesota. Association for Computational Linguistics.

\bibitem[{Vinyals et~al.(2015)Vinyals, Kaiser, Koo, Petrov, Sutskever, and
  Hinton}]{vinyals-etal-2015-grammar}
Oriol Vinyals, \L{}ukasz Kaiser, Terry Koo, Slav Petrov, Ilya Sutskever, and
  Geoffrey Hinton. 2015.
\newblock \href
  {http://papers.nips.cc/paper/5635-grammar-as-a-foreign-language.pdf} {Grammar
  as a {{Foreign Language}}}.
\newblock In \emph{Advances in {{Neural Information Processing Systems}} 28},
  pages 2755--2763. {Curran Associates, Inc.}

\bibitem[{Younger(1967)}]{younger-1967-recognition}
Daniel~H. Younger. 1967.
\newblock Recognition and parsing of context-free languages in time n3.
\newblock \emph{Information and control}, 10(2):189--208.

\bibitem[{Zhang and Clark(2009)}]{zhang-clark-2009-transition}
Yue Zhang and Stephen Clark. 2009.
\newblock \href {https://www.aclweb.org/anthology/W09-3825} {Transition-based
  parsing of the {C}hinese treebank using a global discriminative model}.
\newblock In \emph{Proceedings of the 11th International Conference on Parsing
  Technologies ({IWPT}{'}09)}, pages 162--171, Paris, France. Association for
  Computational Linguistics.

\bibitem[{Zhou and Zhao(2019)}]{zhou-zhao-2019-head}
Junru Zhou and Hai Zhao. 2019.
\newblock \href {https://doi.org/10.18653/v1/P19-1230} {Head-driven phrase
  structure grammar parsing on {P}enn treebank}.
\newblock In \emph{Proceedings of the 57th Annual Meeting of the Association
  for Computational Linguistics}, pages 2396--2408, Florence, Italy.
  Association for Computational Linguistics.

\bibitem[{Zhu et~al.(2013)Zhu, Zhang, Chen, Zhang, and
  Zhu}]{zhu-etal-2013-fast}
Muhua Zhu, Yue Zhang, Wenliang Chen, Min Zhang, and Jingbo Zhu. 2013.
\newblock \href {https://www.aclweb.org/anthology/P13-1043} {Fast and accurate
  shift-reduce constituent parsing}.
\newblock In \emph{Proceedings of the 51st Annual Meeting of the Association
  for Computational Linguistics (Volume 1: Long Papers)}, pages 434--443,
  Sofia, Bulgaria. Association for Computational Linguistics.

\end{thebibliography}
\bibliographystyle{acl_natbib}

\end{document}